\documentclass[10pt,twocolumn,letterpaper]{article}

\usepackage{iccv}
\usepackage{times}
\usepackage{epsfig}
\usepackage{graphicx}
\usepackage{amsmath}
\usepackage{amssymb}
\usepackage{booktabs}
\usepackage[final]{microtype}
\usepackage{bbm}

\usepackage[pagebackref=true,breaklinks=true,letterpaper=true,colorlinks,bookmarks=false]{hyperref}

\iccvfinalcopy 

\def\iccvPaperID{937} 

\newcommand{\pR}{\theta_{\text{repr}}}
\newcommand{\pC}{\theta_{C}}
\newcommand{\pD}{\theta_D}
\newcommand{\lSoft}{\mathcal{L}_{\text{soft}}}
\newcommand{\lConf}{\mathcal{L}_{\text{conf}}}
\newcommand{\lC}{\mathcal{L}_C}
\newcommand{\lD}{\mathcal{L}_D}
\newcommand{\hConf}{\lambda}
\newcommand{\hSoft}{\nu}
\newcommand{\temp}{\tau}

\newcommand\fakefootnote[1]{%
  \begingroup
  \renewcommand\thefootnote{}\footnote{#1}%
  \addtocounter{footnote}{-1}%
  \endgroup
}

\ificcvfinal\fi
\begin{document}

\title{Simultaneous Deep Transfer Across Domains and Tasks}
\author{
Eric Tzeng$^*$, Judy Hoffman$^*$, Trevor Darrell \\
UC Berkeley, EECS \& ICSI \\
{\tt\small \{etzeng,jhoffman,trevor\}@eecs.berkeley.edu}
\and
Kate Saenko \\
UMass Lowell, CS \\
{\tt\small saenko@cs.uml.edu}
}

\maketitle
\fakefootnote{$^*$ Authors contributed equally.}
\vspace{-1em}

\begin{abstract}
  Recent reports suggest that a generic supervised deep CNN model trained on a
  large-scale dataset reduces, but does not remove, dataset bias.
  Fine-tuning deep models in a new domain can require a significant
  amount of labeled data, which for many applications is simply not available.  We
  propose a new CNN architecture to exploit unlabeled and sparsely labeled target domain data.
  Our approach simultaneously optimizes for domain invariance to facilitate domain transfer and uses a soft label distribution matching loss to transfer information between tasks.
  Our proposed adaptation method
  offers empirical performance which exceeds previously published results on two
  standard benchmark visual domain adaptation tasks, evaluated across supervised and {semi-supervised} adaptation settings.
\end{abstract}

\section{Introduction}
Consider a group of robots trained by the manufacturer to recognize thousands of common objects using standard image databases, then shipped to households around the country. As each robot starts to operate in its own unique environment, it is likely to have degraded performance due to the shift in domain. 
It is clear that, given enough extra supervised data from the new environment, the original performance could be recovered. 
However, state-of-the-art recognition algorithms rely on high capacity convolutional neural network (CNN) models that require millions of supervised images for initial training. Even the traditional approach for adapting deep models, fine-tuning~\cite{rcnn, overfeat}, may require hundreds or thousands of labeled examples for each object category that needs to be adapted.

\begin{figure}
\centering
\includegraphics[width=0.9\linewidth]{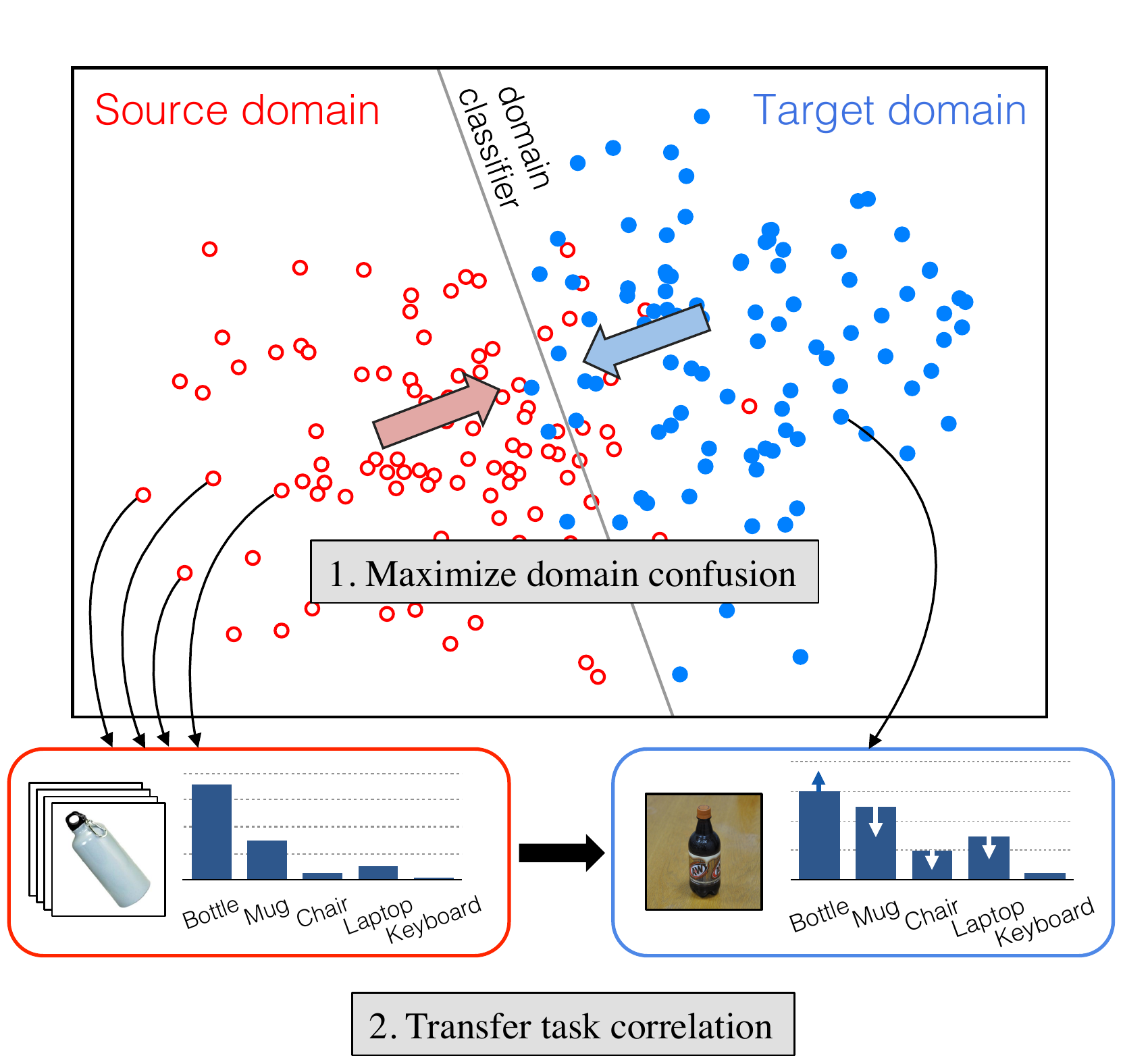}
\vspace{0em} 
\caption{We transfer discriminative category information from a source domain to a target domain via two methods.
First, we maximize domain confusion by making the marginal distributions of the two domains as similar as possible.
Second, we transfer correlations between classes learned on the source examples directly to the target examples, thereby preserving the relationships between classes.}
\label{fig:fig1}
\end{figure}

It is reasonable to assume that the robot's new owner will label a handful of examples for a few types of objects, but completely unrealistic to presume full supervision in the new environment. Therefore, we propose an algorithm that
effectively adapts between the training (source) and test (target) environments by
 utilizing both generic statistics from unlabeled data collected in the new environment as well as a few human labeled examples from a subset of the categories of interest.
Our approach performs transfer learning both across domains and across tasks (see Figure~\ref{fig:fig1}).
Intuitively, domain transfer is accomplished by making the marginal feature distributions of source and target as similar to each other as possible. Task transfer is enabled by transferring empirical category correlations learned on the source to the target domain. This helps to preserve relationships between categories, e.g., \textit{bottle} is similar to \textit{mug} but different from \textit{keyboard}.
Previous work proposed techniques for domain transfer with CNN models~\cite{ganin2015, long2015dan} but did not utilize the learned source semantic structure for task transfer.

To enable domain transfer, 
we use the unlabeled target data to compute an estimated marginal distribution over the new environment and explicitly optimize a feature representation that minimizes the distance between the source and target domain distributions.
Dataset bias was classically illustrated in computer vision by the ``name the dataset'' game of Torralba and Efros~\cite{efros-cvpr11}, which trained a classifier to predict which dataset an image originates from, thereby showing that visual datasets are biased samples of the visual world.
Indeed, this turns
out to be formally connected to measures of domain discrepancy~\cite{adist,mmd}.
Optimizing for domain invariance, therefore, can be considered equivalent to the
task of learning to predict the class labels while simultaneously finding a
representation that makes the domains appear as similar as possible. This
principle forms the 
domain transfer component
of our proposed approach. We learn deep representations
by optimizing over a loss which includes both classification error on the
labeled data as well as a \emph{domain confusion} loss which seeks to make the
domains indistinguishable.

However, while maximizing domain confusion pulls the marginal distributions of the domains together, it does not necessarily align the classes in the target with those in the source.
Thus, we also explicitly transfer the similarity structure amongst categories from the source to the target and further optimize our representation to produce the same structure in the target domain using the few target labeled examples as reference points.
We are inspired by prior work on distilling deep models~\cite{ba-nips14,hinton-nips14} and extend the ideas presented in these works to a domain adaptation setting.
We first compute the average output probability distribution, or ``soft label,'' over the source training examples in each category. Then, for each target labeled example, we directly optimize our model to match the distribution over classes to the soft label. In this way we are able to perform task adaptation by transferring information to categories with no explicit labels in the target domain.

We solve the two problems jointly using a new CNN architecture,
 outlined in Figure~\ref{fig:architecture}.
We combine a domain confusion and softmax cross-entropy losses to train the network with the target data.
%
Our architecture can be used to solve \emph{supervised adaptation}, when a
small amount of target labeled data is available from each category, and
\emph{semi-supervised adaptation}, when a small amount of target labeled data is available from a subset of the categories.
We provide a
comprehensive evaluation on the popular Office benchmark~\cite{saenko-eccv10} and the recently introduced cross-dataset collection~\cite{crossdata} for classification
across visually distinct domains. We demonstrate that by
jointly optimizing for domain confusion and matching soft labels, we are able to
 outperform the current state-of-the-art visual domain adaptation
results.
%

\section{Related work}
There have been many approaches proposed in recent years to solve the visual domain
adaptation problem, which is also commonly framed as the visual dataset bias problem~\cite{efros-cvpr11}.
All recognize that there is a shift in the distribution of
the source and target data representations. In fact, the size of a domain shift
is often measured by the distance between the source and target subspace
representations~\cite{mmd,sa,adist, disc,tca}. A large number of methods have
sought to overcome this difference by learning a feature space transformation to
align the source and target representations~\cite{saenko-eccv10,kulis-cvpr11,
sa, gong-cvpr12}. For the \emph{supervised} adaptation scenario, when a limited
amount of labeled data is available in the target domain, some approaches have
been proposed to learn a target classifier regularized against the source
classifier~\cite{yang-icdm07, aytar-iccv11, BergamoTorresani10}. Others have
sought to both learn a feature transformation and regularize a target classifier
simultaneously~\cite{hoffman-iclr13, duan-icml12}.

Recently, supervised CNN based feature
representations have been shown to be extremely effective for a variety of
visual recognition tasks~\cite{supervision,decaf, rcnn, overfeat}.  In
particular, using deep representations dramatically reduces the effect of
resolution and lighting on domain shifts~\cite{decaf, hoffman-iclr14}.
Parallel CNN architectures such as Siamese networks have been shown to be
effective for learning invariant representations~\cite{bromley1993signature,
chopra2005learning}. However, training these networks requires labels for each
training instance, so it is unclear how to extend these methods to unsupervised
or semi-supervised settings.
Multimodal deep learning architectures have also been explored to learn
representations that are invariant to different input
modalities~\cite{ngiam2011multimodal}. However, this method operated primarily
in a generative context and therefore did not leverage the full representational
power of supervised CNN representations.

Training a joint source and target CNN architecture was proposed
by~\cite{ref:dlid}, but was limited to two layers and so was significantly
outperformed by the methods which used a deeper architecture~\cite{supervision},
pre-trained on a large auxiliary data source (ex: ImageNet~\cite{ilsvrc2012}).
\cite{da-mmd} proposed pre-training with a denoising auto-encoder, then training a two-layer network
simultaneously with the MMD domain confusion loss. This effectively learns a
domain invariant representation, but again, because the learned network is
relatively shallow, it lacks the strong semantic representation that is learned
by directly optimizing a classification objective with a supervised deep CNN.

Using classifier output distributions instead of category labels during training
has been explored in the context of model compression or distillation~\cite{ba-nips14,hinton-nips14}.
However, we are the first to apply this technique in a domain adaptation setting
in order to transfer class correlations between domains.

Other works have cotemporaneously explored the idea of directly optimizing a representation for domain invariance~\cite{ganin2015,long2015dan}.
However, they either use weaker measures of domain invariance or make use of optimization methods that are less robust than our proposed method, and they do not attempt to solve the task transfer problem in the semi-supervised setting.

\section{Joint CNN architecture for domain and task transfer}
\begin{figure*}
\centering
\includegraphics[width=.8\linewidth]{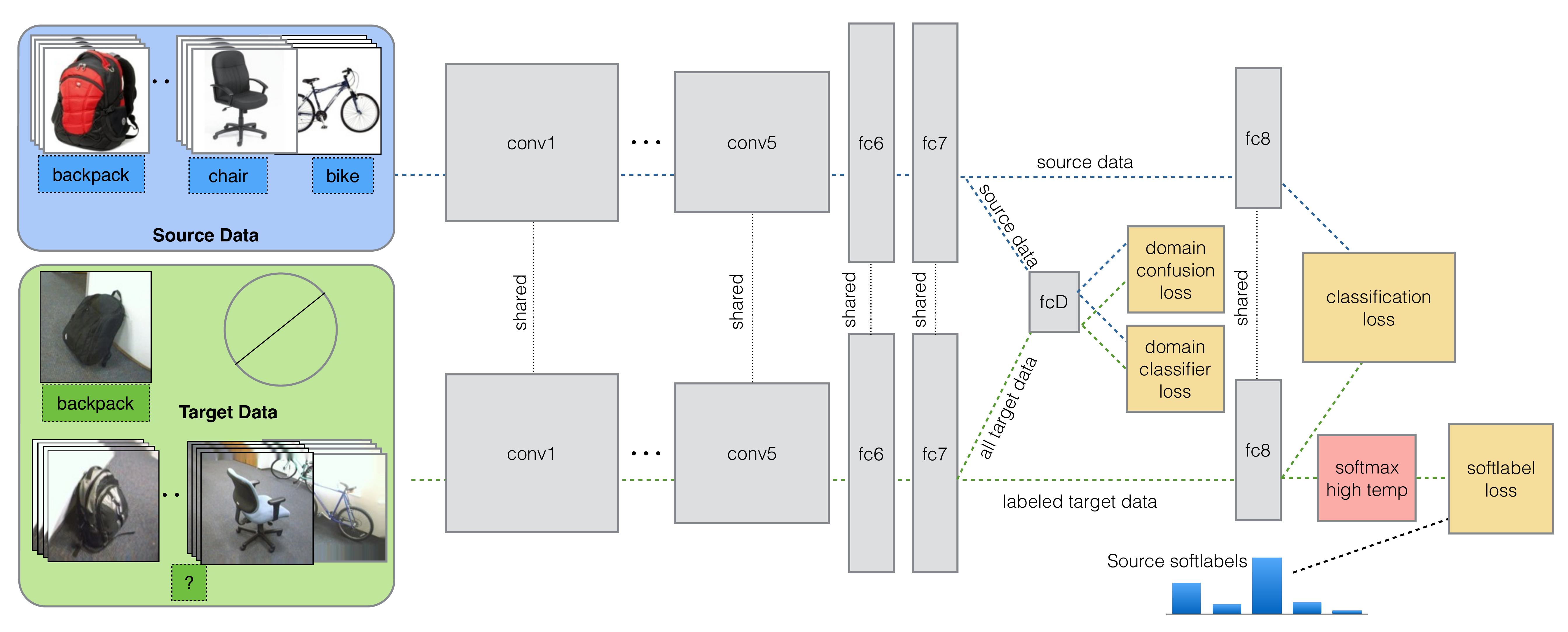}
\caption{Our overall CNN architecture for domain and task transfer. We use a domain confusion loss over all source and target (both labeled and unlabeled) data to learn a domain invariant representation. We simultaneously transfer the learned source semantic structure to the target domain by optimizing the network to produce activation distributions that match those learned for source data in the source only CNN. \emph{Best viewed in color.}}

\label{fig:architecture}
\end{figure*}

We first give an overview of our convolutional network (CNN) architecture, depicted in Figure~\ref{fig:architecture}, that learns a representation which both aligns visual domains and transfers the semantic structure from a well labeled source domain to the sparsely labeled target domain.
We assume access to a limited amount of labeled target data, potentially from only a subset of the categories of interest. With limited labels on a subset of the categories, the traditional domain transfer approach of fine-tuning on the available target data~\cite{rcnn,overfeat,lsda} is not effective. Instead, since the source labeled data shares the label space of our target domain, we use the source data to guide training of the corresponding classifiers.


Our method takes as input the labeled source data $\{x_S, y_S\}$ (blue box Figure~\ref{fig:architecture}) and the target data $\{x_T, y_T\}$ (green box Figure~\ref{fig:architecture}), where the labels $y_T$ are only provided for a subset of the target examples.
Our goal is to produce a category classifier $\pC$ that operates on an image feature representation $f(x; \pR)$ parameterized by representation parameters $\pR$ and can correctly classify target examples at test time.

For a setting with $K$ categories, let our desired classification objective be defined as the standard softmax loss
\begin{equation}
  \lC(x,y; \pR, \pC) = - \sum_k \mathbbm{1}[y=k]\log p_k \label{eq:obj-c}
\end{equation}
where $p$ is the softmax of the classifier activations, $p =\nobreak\text{softmax}(\pC^T f(x; \pR))$.

We could use the available source labeled data to train our representation and classifier parameters according to Equation~\eqref{eq:obj-c}, but this often leads to overfitting to the source distribution, causing reduced performance at test time when recognizing in the target domain. 
However, we note that if the source and target domains are very similar then the classifier trained on the source will perform well on the target. In fact, 
it is sufficient
for the source and target data to be similar under the learned representation, $\pR$. 

Inspired by the ``name the dataset'' game of Torralba and Efros~\cite{efros-cvpr11}, we can directly train a domain classifier $\pD$ to identify whether a training example originates from the source or target domain given its feature representation.
Intuitively, if our choice of representation suffers from domain shift, then they will lie in distinct parts of the feature space, and a classifier will be able to easily separate the domains.
We use this notion to add a new \emph{domain confusion} loss $\lConf(x_S, x_T, \pD; \pR)$ to our objective and directly optimize our representation so as to minimize the discrepancy between the source and target distributions.
This loss is described in more detail in Section~\ref{sec:method-confusion}.

Domain confusion can be applied to learn a representation that aligns source and target data without any target labeled data. However, we also presume a handful of sparse labels in the target domain, $y_T$.
In this setting, a simple approach is to incorporate the target labeled data along with the source labeled data into the classification objective of Equation~\eqref{eq:obj-c}\footnote{We present this approach as one of our baselines.}. However, fine-tuning with hard category labels limits the impact of a single training example, making it hard for the network to learn to generalize from the limited labeled data.
Additionally, fine-tuning with hard labels is ineffective when labeled data is available for only a subset of the categories.

For our approach, we draw inspiration from recent network distillation works~\cite{ba-nips14, hinton-nips14}, which demonstrate that a large network can be ``distilled'' into a simpler model by replacing the hard labels with the softmax activations from the original large model.
This modification proves to be critical, as the distribution holds key information about the relationships between categories and imposes additional structure during the training process.
In essence, because each training example is paired with an output distribution, it provides valuable information about not only the category it belongs to, but also each other category the classifier is trained to recognize.

Thus, we propose using the labeled target data to optimize the network parameters through a \emph{soft label} loss, $\lSoft(x_T, y_T ; \pR, \pC)$. This loss will train the network parameters to produce a ``soft label'' activation that matches the average output distribution of source examples on a network trained to classify source data. This loss is described in more detail in Section~\ref{sec:method-softlabels}.
By training the network to match the expected source output distributions on target data, we transfer the learned inter-class correlations from the source domain to examples in the target domain.
This directly transfers useful information from source to target, such as the fact that \emph{bookshelves} appear more similar to \emph{filing cabinets} than to \emph{bicycles}.


Our full method then minimizes the joint loss function
\begin{equation}
\begin{split}
\mathcal{L}(x_S, y_S, x_T, y_T, \pD ; &\pR, \pC) = \\
  &\lC(x_S, y_S, x_T, y_T; \pR, \pC) \\ 
  &+ \hConf \lConf(x_S, x_T, \pD; \pR) \\ 
  &+ \hSoft \lSoft(x_T, y_T ; \pR, \pC).
\end{split}
\end{equation}
where the hyperparameters $\hConf$ and $\hSoft$ determine how strongly domain confusion and soft labels influence the optimization.


Our ideas of domain confusion and soft label loss for task transfer are generic and can be applied to any CNN classification architecture. For our experiments and for the detailed discussion in this paper we modify the standard Krizhevsky architecture~\cite{supervision}, which has five convolutional layers (conv1--conv5) and three fully connected layers (fc6--fc8). The representation parameter $\pR$ corresponds to layers 1--7 of the network, and the classification parameter $\pC$ corresponds to layer 8. For the remainder of this section, we provide further details on our novel loss definitions and the implementation of our model.

\subsection{Aligning domains via domain confusion}
\label{sec:method-confusion}
In this section we describe in detail our proposed \emph{domain confusion} loss objective. 
Recall that we introduce the domain confusion loss as a means to learn a representation that is domain invariant, and thus will allow us to better utilize a classifier trained using the labeled source data.
We consider a representation to be domain invariant if a classifier trained using that representation can not distinguish examples from the two domains.

To this end, we add an additional domain classification layer, denoted as fcD in Figure~\ref{fig:architecture}, with parameters $\pD$. This layer simply performs binary classification using the domain corresponding to an image as its label.
For a particular feature representation, $\pR$, we evaluate its domain invariance by learning the best domain classifier on the representation.
This can be learned by optimizing the following objective, where $y_D$ denotes the domain that the example is drawn from:
\begin{equation}
  \lD(x_S, x_T, \pR; \pD) = -\sum_d \mathbbm{1}[y_D=d] \log q_d \label{eq:obj-dom-classifier}
\end{equation}
with $q$ corresponding to the softmax of the domain classifier activation: $q = \text{softmax}(\pD^T f(x; \pR))$.

For a particular domain classifier, $\pD$, we can now introduce our loss which seeks to ``maximally confuse'' the two domains by computing the cross entropy between the output predicted domain labels and a uniform distribution over domain labels:
\begin{equation}
  \phantom{.}\lConf(x_S, x_T, \pD; \pR) = -\sum_d \frac{1}{D} \log q_d. \label{eq:obj-confusion}
\end{equation}
This domain confusion loss seeks to learn domain invariance by finding a representation in which the best domain classifier performs poorly.

Ideally, we want to simultaneously minimize Equations~\eqref{eq:obj-dom-classifier} and~\eqref{eq:obj-confusion} for the representation and the domain classifier parameters.
However, the two losses stand in direct opposition to one another: learning a fully domain invariant representation means the domain classifier must do poorly, and learning an effective domain classifier means that the representation is not domain invariant.
Rather than globally optimizing $\pD$ and $\pR$, we instead perform
iterative updates for the following two objectives given the fixed parameters from the previous iteration:
\begin{align}
\min_{\pD}\ &\lD(x_S, x_T, \pR; \pD) \label{eq:min-dom} \\
\min_{\pR}\ &\lConf(x_S, x_T, \pD; \pR). \label{eq:min-conf}
\end{align}

These losses are readily implemented in standard deep learning frameworks, and after setting learning rates properly so that Equation~\eqref{eq:min-dom} only updates $\pD$ and Equation~\eqref{eq:min-conf} only updates $\pR$, the updates can be performed via standard backpropagation.
Together, these updates ensure that we learn a representation that is domain invariant.

\subsection{Aligning source and target classes via soft labels}
\label{sec:method-softlabels}

\begin{figure}
\centering
\includegraphics[width = \linewidth]{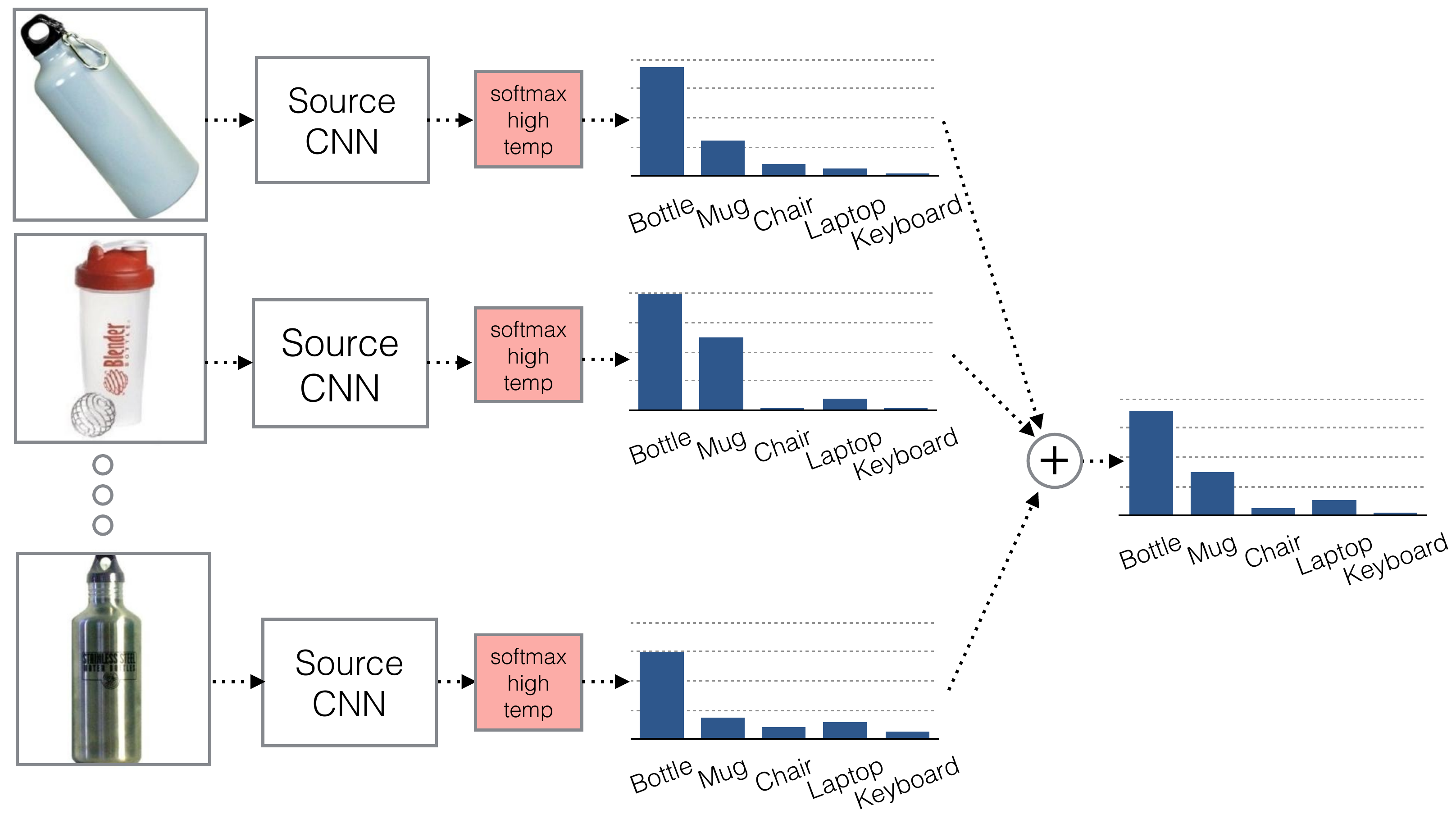}
\caption{Soft label distributions are learned by averaging the per-category activations of source training examples using the source model. An example, with 5 categories, depicted here to demonstrate the final soft activation for the bottle category will be primarily dominated by bottle and mug with very little mass on chair, laptop, and keyboard. }
\label{fig:soft-src}
\end{figure}

\begin{figure}
\centering
\includegraphics[width = \linewidth]{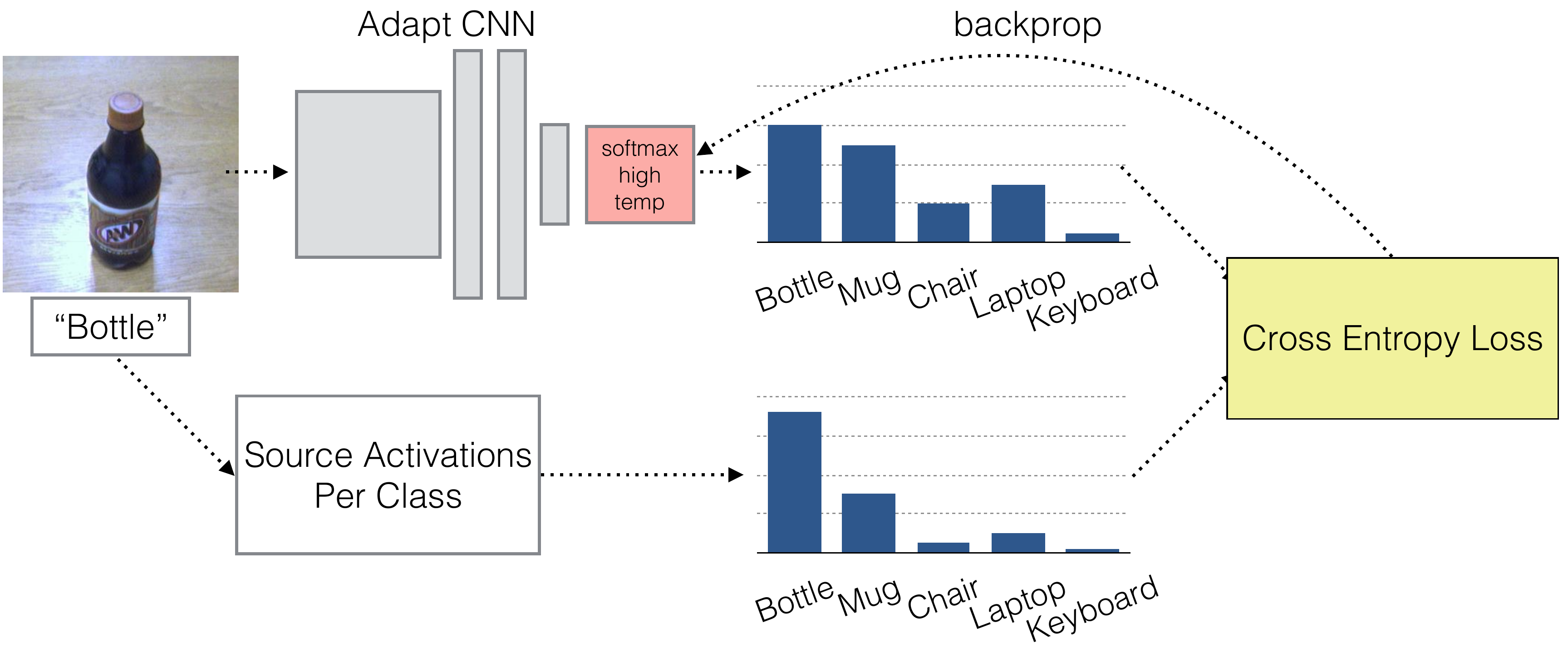}
\caption{Depiction of the use of source per-category soft activations with the cross entropy loss function over the current target activations. }
\label{fig:soft-crossent}
\end{figure}

While training the network to confuse the domains acts to align their marginal distributions, there are no guarantees about the alignment of classes between each domain.
To ensure that the relationships between classes are preserved across source and target, we fine-tune the network against ``soft labels'' rather than the image category hard label.



We define a soft label for category $k$ as the average over the softmax of all activations of source examples in category $k$, depicted graphically in Figure~\ref{fig:soft-src}, and denote this average as $l^{(k)}$. Note that, since the source network was trained purely to optimize a classification objective, a simple softmax over each $z^i_S$ will hide much of the useful information by producing a very peaked distribution. Instead, we use a softmax with a high temperature $\temp$ so that the related classes have enough probability mass to have an effect during fine-tuning. 
With our computed per-category soft labels we can now 
define our soft label loss:
\begin{equation}
  \lSoft(x_T,y_T; \pR, \pC) = -\sum_i l^{(y_T)}_i \log p_i 
\end{equation}
where $p$ denotes the soft activation of the target image, $p=\text{softmax}(\pC^T f(x_T;\pR )/\temp)$. 
The loss above corresponds to the cross-entropy loss between the soft activation of a particular target image and the soft label corresponding to the category of that image, as shown in Figure~\ref{fig:soft-crossent}.


To see why this will help, consider the soft label for a particular category, such as \emph{bottle}. 
The soft label $l^{(\text{bottle})}$ is a $K$-dimensional vector, where each dimension indicates the similarity of bottles to each of the $K$ categories. 
In this example, the bottle soft label will have a higher weight on \emph{mug} than on \emph{keyboard}, since bottles and mugs are more visually similar.
Thus, soft label training with this particular soft label directly enforces the relationship that bottles and mugs should be closer in feature space than bottles and keyboards.

One important benefit of using this soft label loss is that we ensure that the parameters for categories without any labeled target data are still updated to output non-zero probabilities.
We explore this benefit in Section~\ref{sec:eval}, where we train a network using labels from a subset of the target categories and find significant performance improvement even when evaluating only on the unlabeled categories. 



\section{Evaluation}
\label{sec:eval}
To analyze the effectiveness of our method, we evaluate it on the Office dataset, a standard benchmark dataset for visual domain adaptation, and on a new large-scale cross-dataset domain adaptation challenge.

\subsection{Adaptation on the Office dataset}
\begin{table*}
  \setlength{\tabcolsep}{4pt}
\centering
\begin{tabular}{lcccccc|c}
\toprule
                                 & $A \rightarrow W$           & $A \rightarrow D$           & $W \rightarrow A$           & $W \rightarrow D$        & $D \rightarrow A$           & $D \rightarrow W$           & Average  \\
\midrule
DLID~\cite{ref:dlid}             &            51.9             & --                          & --                          &         89.9             & --                          &            78.2             & -- \\
DeCAF$_6$ S+T~\cite{decaf}       &            80.7 $\pm$ 2.3   & --                          & --                          & --                       & --                          &            94.8 $\pm$ 1.2   & -- \\
DaNN~\cite{da-mmd}               &            53.6 $\pm$ 0.2   & --                          & --                          &         83.5 $\pm$ 0.0   & --                          &            71.2 $\pm$ 0.0   & -- \\
Source CNN                       &            56.5 $\pm$ 0.3   &            64.6 $\pm$ 0.4   &            42.7 $\pm$ 0.1   &         93.6 $\pm$ 0.2   &            47.6 $\pm$ 0.1   &            92.4 $\pm$ 0.3   &         66.22 \\
Target CNN                       &            80.5 $\pm$ 0.5   &            81.8 $\pm$ 1.0   &            59.9 $\pm$ 0.3   &         81.8 $\pm$ 1.0   &            59.9 $\pm$ 0.3   &            80.5 $\pm$ 0.5   &         74.05 \\
Source+Target CNN                & \underline{82.5 $\pm$ 0.9}  & \underline{85.2 $\pm$ 1.1}  &    \textbf{65.2 $\pm$ 0.7}  &         96.3 $\pm$ 0.5   & \underline{65.8 $\pm$ 0.5}  &            93.9 $\pm$ 0.5   &         81.50 \\
\midrule
Ours: dom confusion only         &    \textbf{82.8 $\pm$ 0.9}  & \underline{85.9 $\pm$ 1.1}  & \underline{64.9 $\pm$ 0.5}  &         97.5 $\pm$ 0.2   &    \textbf{66.2 $\pm$ 0.4}  & \underline{95.6 $\pm$ 0.4}  &         82.13 \\
Ours: soft labels only           & \underline{82.7 $\pm$ 0.7}  &            84.9 $\pm$ 1.2   &    \textbf{65.2 $\pm$ 0.6}  & \textbf{98.3 $\pm$ 0.3}  & \underline{66.0 $\pm$ 0.5}  &    \textbf{95.9 $\pm$ 0.6}  &         82.17 \\
Ours: dom confusion+soft labels  & \underline{82.7 $\pm$ 0.8}  &    \textbf{86.1 $\pm$ 1.2}  & \underline{65.0 $\pm$ 0.5}  &         97.6 $\pm$ 0.2   &    \textbf{66.2 $\pm$ 0.3}  & \underline{95.7 $\pm$ 0.5}  & \textbf{82.22}\\
\bottomrule
\end{tabular}

\caption{Multi-class accuracy evaluation on the standard supervised adaptation
  setting with the \emph{Office} dataset.  We evaluate on all 31 categories
  using the standard experimental protocol from \cite{saenko-eccv10}. Here, we
  compare against three state-of-the-art domain adaptation methods as well as a CNN trained using only source data, only target data, or both source and target data together.}
\label{table:full-super}
\end{table*}

The Office dataset is a collection of images from three distinct domains, Amazon, DSLR, and Webcam, the largest of which has 2817 labeled images~\cite{saenko-eccv10}.
The 31 categories in the dataset consist of objects commonly encountered in office settings, such as keyboards, file cabinets, and laptops.

We evaluate our method in two different settings:
\begin{itemize} \itemsep1pt \parskip0pt \parsep0pt
  \item \textbf{Supervised adaptation} Labeled training data for all categories is available in source and sparsely in target.
  \item \textbf{Semi-supervised adaptation (task adaptation)} Labeled training data is available in source and sparsely for a subset of the target categories.
\end{itemize}

\begin{table*}
  \setlength{\tabcolsep}{4pt}
\centering
\begin{tabular}{lcccccc|c}
\toprule
                                & $A \rightarrow W$        & $A \rightarrow D$        & $W \rightarrow A$        & $W \rightarrow D$        & $D \rightarrow A$        & $D \rightarrow W$       & Average  \\
\midrule
MMDT~\cite{hoffman-iclr13}      & --                       &         44.6 $\pm$ 0.3   & --                       &         58.3 $\pm$ 0.5   & --                       & --                      & --   \\
Source CNN                      &         54.2 $\pm$ 0.6   &         63.2 $\pm$ 0.4   &         34.7 $\pm$ 0.1   &         94.5 $\pm$ 0.2   &         36.4 $\pm$ 0.1   &         89.3 $\pm$ 0.5  &         62.0 \\
\midrule
Ours: dom confusion only        &         55.2 $\pm$ 0.6   &         63.7 $\pm$ 0.9   & \textbf{41.1 $\pm$ 0.0}  &         96.5 $\pm$ 0.1   &         41.2 $\pm$ 0.1   & \textbf{91.3 $\pm$ 0.4} &         64.8 \\
Ours: soft labels only          &         56.8 $\pm$ 0.4   &         65.2 $\pm$ 0.9   &         38.8 $\pm$ 0.4   &         96.5 $\pm$ 0.2   &         41.7 $\pm$ 0.3   &         89.6 $\pm$ 0.1  &         64.8 \\
Ours: dom confusion+soft labels & \textbf{59.3 $\pm$ 0.6}  & \textbf{68.0 $\pm$ 0.5}  &         40.5 $\pm$ 0.2   & \textbf{97.5 $\pm$ 0.1}  & \textbf{43.1 $\pm$ 0.2}  &         90.0 $\pm$ 0.2  & \textbf{66.4}\\
\bottomrule
\end{tabular}

\caption{Multi-class accuracy evaluation on the standard semi-supervised adaptation
  setting with the \emph{Office} dataset.  We evaluate on 16 held-out categories for which we have no access to target labeled data.
We show results on these unsupervised categories for the source only model, our model trained using only soft labels for the 15 auxiliary categories, and finally using domain confusion together with soft labels on the 15 auxiliary categories.}
\label{table:full-semi}
\end{table*}



For all experiments we initialize the parameters of conv1--fc7 using the released CaffeNet~\cite{caffe} weights. 
We then further fine-tune the network using the source labeled data in order to produce the soft label distributions and use the learned source CNN weights as the initial parameters for training our method. All implementations are produced using the open source Caffe~\cite{caffe} framework, and the network definition files and cross entropy loss layer needed for training will be released upon acceptance. We optimize the network using a learning rate of 0.001 and set the hyper-parameters to $\hConf=0.01$ (confusion) and $\hSoft=0.1$ (soft).

For each of the six domain shifts, we evaluate across five train/test splits, which are generated by sampling examples from the full set of images per domain.
In the source domain, we follow the standard protocol for this dataset and generate splits by sampling 20 examples per category for the Amazon domain, and 8 examples per category for the DSLR and Webcam domains.
We first present results for the supervised setting, where 3 labeled examples are provided for each category in the target domain.
We report accuracies on the remaining unlabeled images, following the standard protocol introduced with the dataset~\cite{saenko-eccv10}.
In addition to a variety of baselines, we report numbers for both soft label fine-tuning alone as well as soft labels with domain confusion in Table~\ref{table:full-super}.
Because the Office dataset is imbalanced, we report multi-class accuracies, which are obtained by computing per-class accuracies independently, then averaging over all 31 categories.

We see that fine-tuning with soft labels or domain confusion provides a consistent improvement over hard label training in 5 of 6 shifts.
Combining soft labels with domain confusion produces marginally higher performance on average.
This result follows the intuitive notion that when enough target labeled examples are present, directly optimizing for the joint source and target classification objective (Source+Target CNN) is a strong baseline and so using either of our new losses adds enough regularization to improve performance.

Next, we experiment with the semi-supervised adaptation setting. We consider the case in which training data and labels are available for some, but not all of the categories in the target domain.
We are interested in seeing whether we can transfer information learned from the labeled classes to the unlabeled classes. 

To do this, we consider having 10 target labeled examples per category from only 15 of the 31 total categories, following the standard protocol introduced with the \emph{Office} dataset~\cite{saenko-eccv10}. We then evaluate our classification performance on the remaining 16 categories for which no data was available at training time. 

In Table~\ref{table:full-semi} we present multi-class accuracies over the 16 held-out categories and compare our method
 to a previous domain adaptation method~\cite{hoffman-iclr13} as well as a source-only trained CNN. Note that, since the performance here is computed over only a subset of the categories in the dataset, the numbers in this table should not be directly compared to the supervised setting in Table~\ref{table:full-super}. 
 
 We find that all variations of our method (only soft label loss, only domain confusion, and both together) outperform the baselines. Contrary to the fully supervised case, here we note that both domain confusion and soft labels contribute significantly to the overall performance improvement of our method. This stems from the fact that we are now evaluating on categories which lack labeled target data, and thus the network can not implicitly enforce domain invariance through the classification objective alone. Separately, the fact that we get improvement from the soft label training on related tasks indicates that information is being effectively transferred between tasks.

\begin{figure}
\centering
\includegraphics[width=1.0\linewidth]{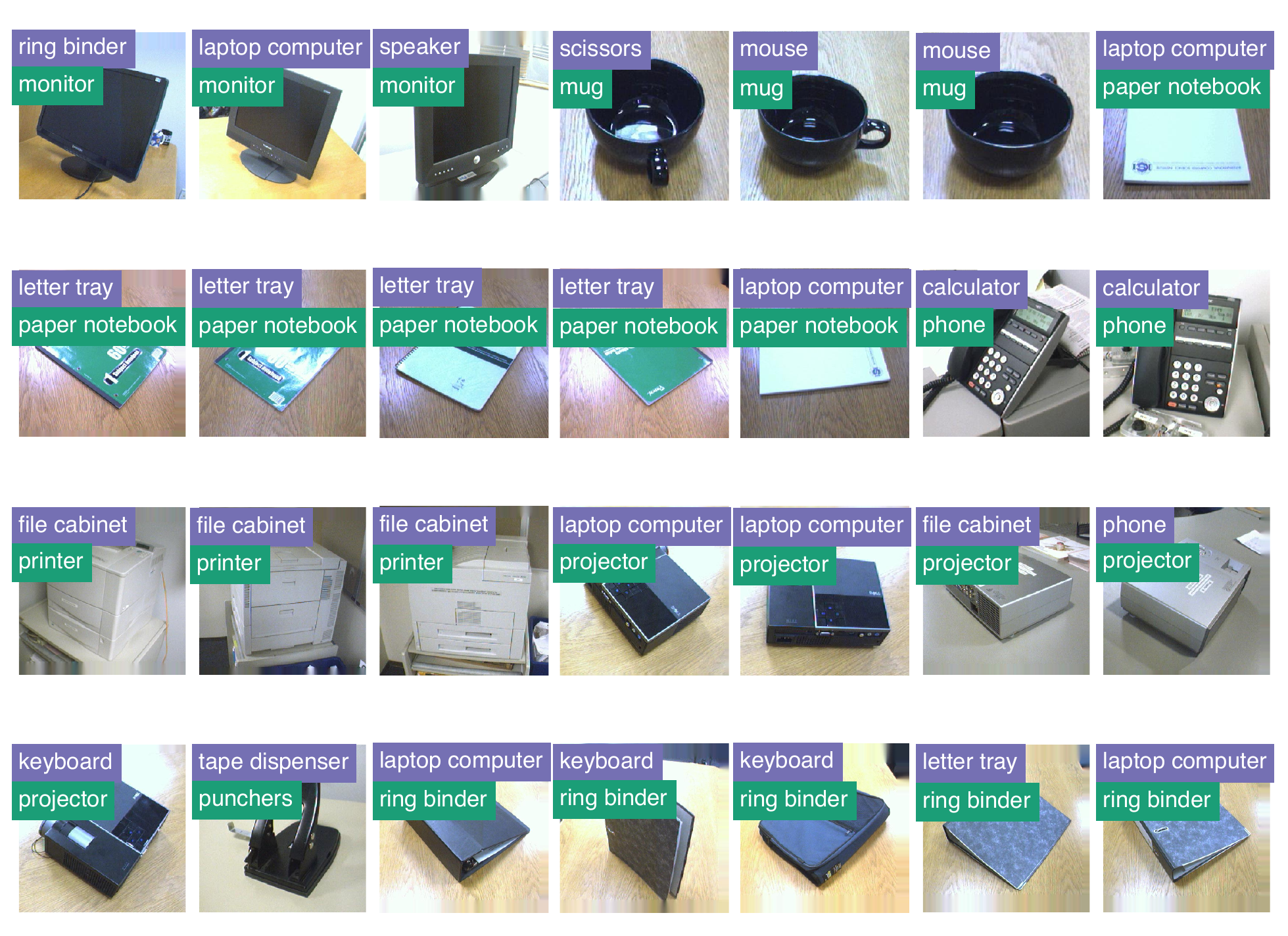}
\caption{Examples from the Amazon$\rightarrow$Webcam shift in the semi-supervised adaptation setting, where our method (the bottom turquoise label) correctly classifies images while the baseline (the top purple label) does not.}
\label{fig:wins_aw}
\end{figure}

 In Figure~\ref{fig:wins_aw}, we show examples for the Amazon$\rightarrow$Webcam shift where our method correctly classifies images from held out object categories and the baseline does not. We find that our method is able to consistently overcome error cases, such as the notebooks that were previously confused with letter trays, or the black mugs that were confused with black computer mice.

\subsection{Adaptation between diverse domains}
For an evaluation with larger, more distinct domains, we test on the recent testbed for cross-dataset analysis~\cite{crossdata}, which collects images from classes shared in common among computer vision datasets.
We use the dense version of this testbed, which consists of 40 categories shared between the ImageNet, Caltech-256, SUN, and Bing datasets, and evaluate specifically with ImageNet as source and Caltech-256 as target.

We follow the protocol outlined in~\cite{crossdata} and generate 5 splits by selecting 5534 images from ImageNet and 4366 images from Caltech-256 across the 40 shared categories.
Each split is then equally divided into a train and test set.
However, since we are most interested in evaluating in the setting with limited target data, we further subsample the target training set into smaller sets with only 1, 3, and 5 labeled examples per category.

\begin{figure}
\centering
\includegraphics[width=.7\linewidth]{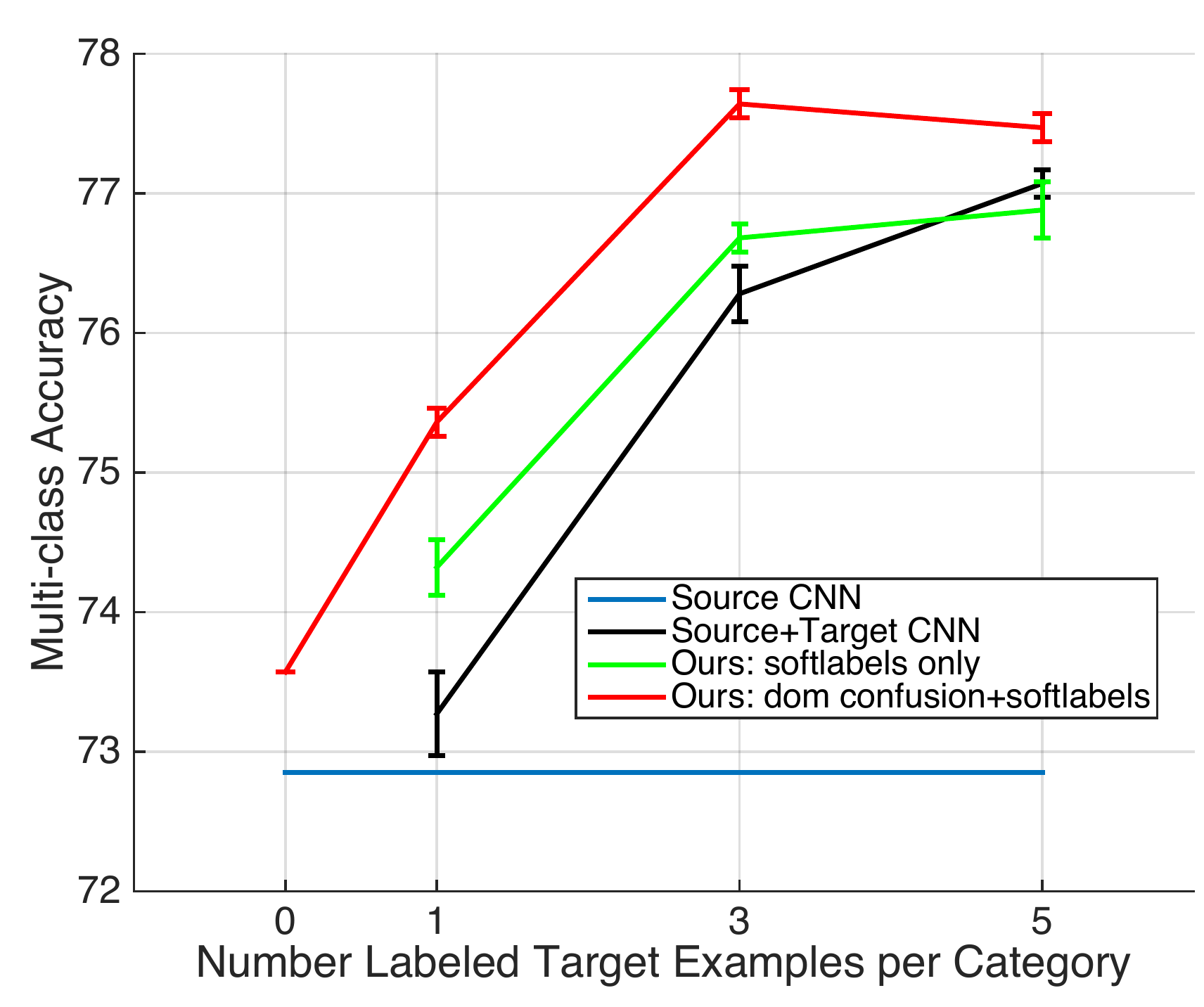}
\caption{ImageNet$\rightarrow$Caltech supervised adaptation from the Cross-dataset~\cite{crossdata} testbed with varying numbers of labeled target examples per category. We find that our method using soft label loss (with and without domain confusion) outperforms the baselines of training on source data alone or using a standard fine-tuning strategy to train with the source and target data. \emph{Best viewed in color.}}
\label{fig:i2c_super}
\end{figure}

Results from this evaluation are shown in Figure~\ref{fig:i2c_super}.
We compare our method to both CNNs fine-tuned using only source data using source and target labeled data.
Contrary to the previous supervised adaptation experiment, our method significantly outperforms both baselines. We see that our full architecture, combining domain confusion with the soft label loss, performs the best overall and is able to operate in the regime of no labeled examples in the target (corresponding to the red line at point 0 on the $x$-axis).
We find that the most benefit of our method arises when there are few labeled training examples per category in the target domain. As we increase the number of labeled examples in the target, the standard fine-tuning strategy begins to approach the performance of the adaptation approach. This indicates that direct joint source and target fine-tuning is a viable adaptation approach when you have a reasonable number of training examples per category. In comparison, fine-tuning on the target examples alone yields accuracies of $36.6 \pm 0.6$, $60.9 \pm 0.5$, and $67.7 \pm 0.5$ for the cases of 1, 3, and 5 labeled examples per category, respectively. All of these numbers underperform the source only model, indicating that adaptation is crucial in the setting of limited training data.

Finally, we note that our results are significantly higher than the $24.8\%$ result reported in~\cite{crossdata}, despite the use of much less training data.
This difference is explained by their use of SURF BoW features, indicating that CNN features are a much stronger feature for use in adaptation tasks.

\section{Analysis}
Our experimental results demonstrate that our method improves classification performance in a variety of domain adaptation settings.
We now perform additional analysis on our method by confirming our claims that it exhibits domain invariance and transfers information across tasks.

\subsection{Domain confusion enforces domain invariance}

\begin{figure}
\centering
\includegraphics[width=\linewidth]{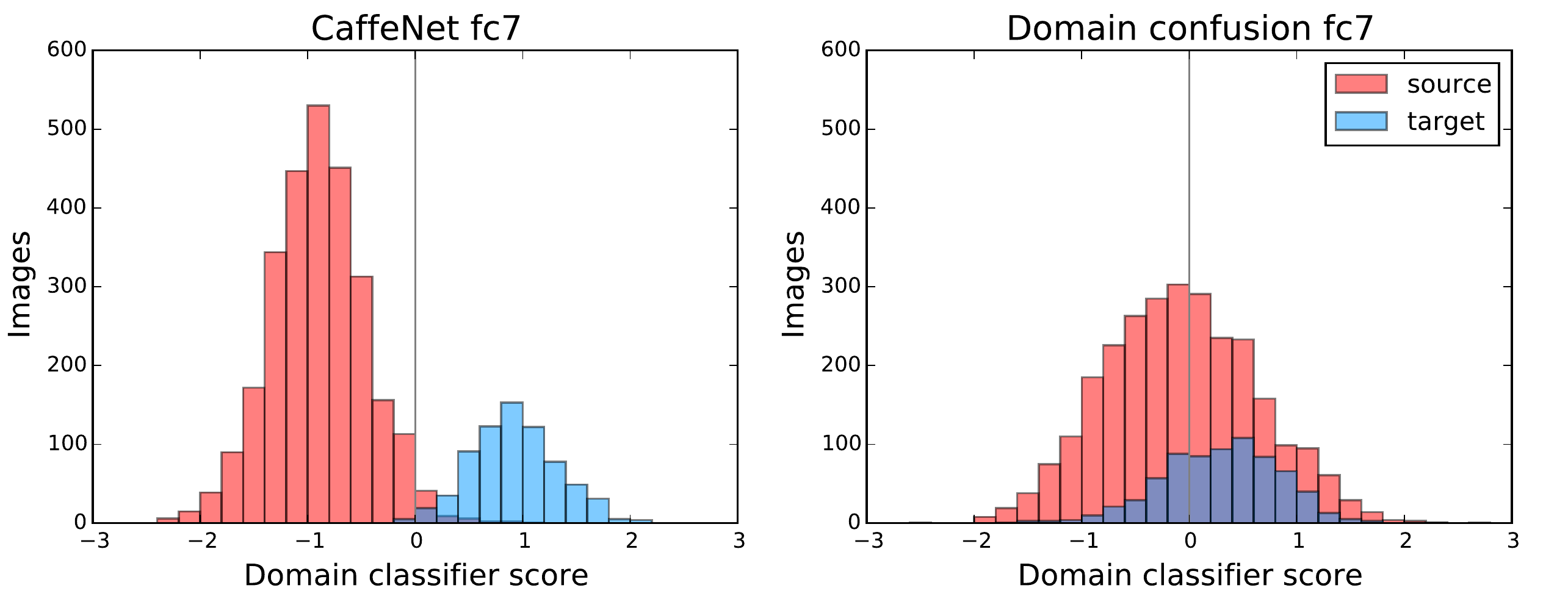}
\caption{We compare the baseline CaffeNet representation to our representation learned with domain confusion by training a support vector machine to predict the domains of Amazon and Webcam images.
  For each representation, we plot a histogram of the classifier decision scores of the test images.
  In the baseline representation, the classifier is able to separate the two domains with 99\% accuracy.
  In contrast, the representation learned with domain confusion is domain invariant, and the classifier can do no better than 56\%.}
\label{fig:qual-confusion}
\end{figure}

We begin by evaluating the effectiveness of domain confusion at learning a domain invariant representation.
As previously explained, we consider a representation to be domain invariant if an optimal classifier has difficulty predicting which domain an image originates from.
Thus, for our representation learned with a domain confusion loss, we expect a trained domain classifier to perform poorly.

We train two support vector machines (SVMs) to classify images into domains: one using the baseline CaffeNet fc7 representation, and the other using our fc7 learned with domain confusion.
These SVMs are trained using 160 images, 80 from Amazon and 80 from Webcam, then tested on the remaining images from those domains.
We plot the classifier scores for each test image in Figure~\ref{fig:qual-confusion}.
It is obvious that the domain confusion representation is domain invariant, making it much harder to separate the two domains---the test accuracy on the domain confusion representation is only 56\%, not much better than random.
In contrast, on the baseline CaffeNet representation, the domain classifier achieves 99\% test accuracy.

\subsection{Soft labels for task transfer}
\begin{figure}
\centering
\includegraphics[width=1.0\linewidth]{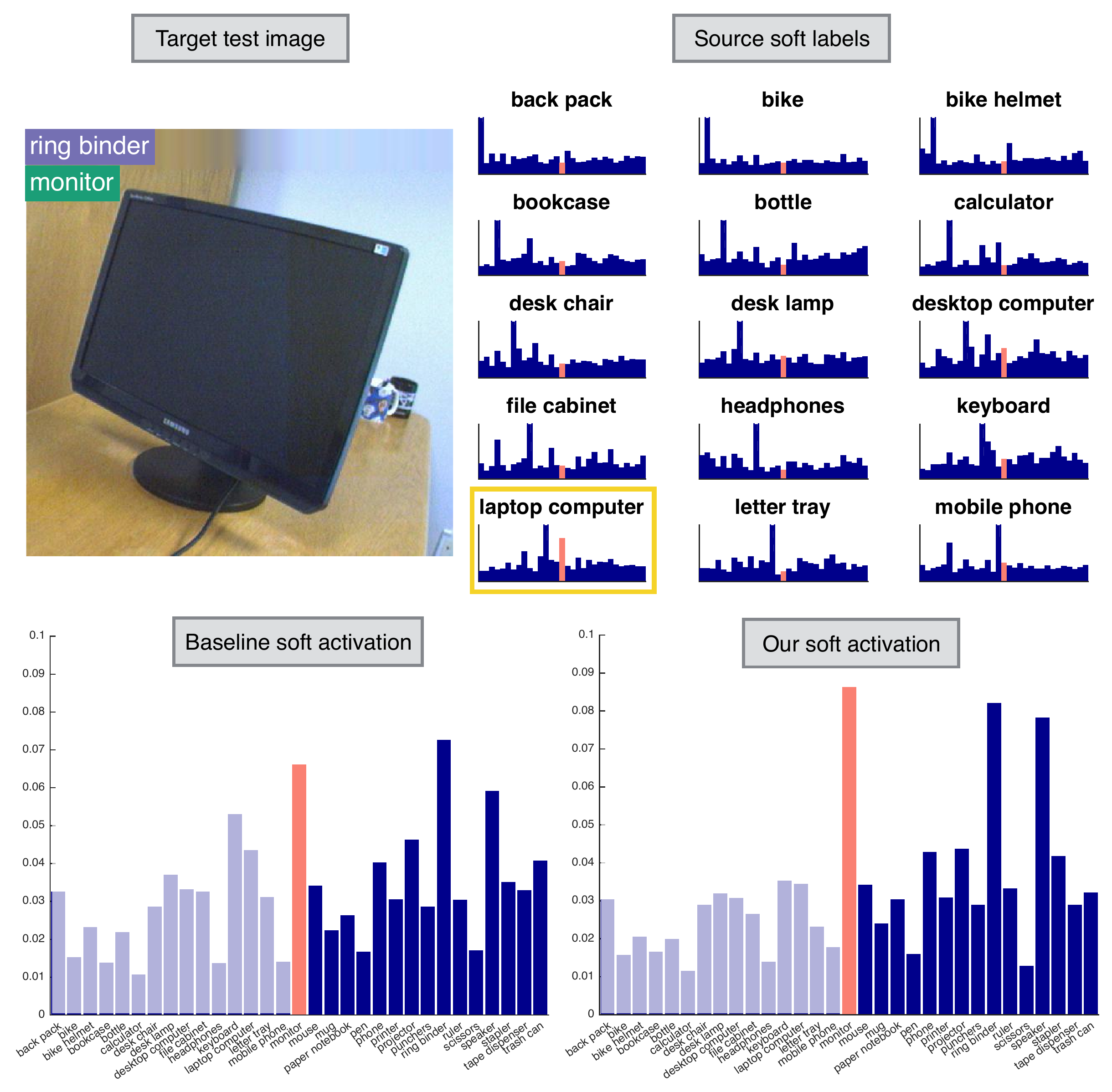}
\caption{Our method (bottom turquoise label) correctly predicts the category of this image, whereas the baseline (top purple label) does not. The source per-category soft labels for the 15 categories with labeled target data are shown in the upper right corner, where the $x$-axis of the plot represents the 31 categories and the $y$-axis is the output probability. We highlight the index corresponding to the \emph{monitor} category in red.  As no labeled target data is available for the correct category, \emph{monitor}, we find that in our method the related category of \emph{laptop computer} (outlined with yellow box) transfers information to the monitor category. As a result, after training, our method places the highest weight on the correct category. Probability score per category for the baseline and our method are shown in the bottom left and right, respectively, training categories are opaque and correct test category is shown in red.}
\label{fig:softlabel-res}
\end{figure}

We now examine the effect of soft labels in transferring information between categories. We consider the Amazon$\rightarrow$Webcam shift from the semi-supervised adaptation experiment in the previous section. Recall that in this setting, we have access to target labeled data for only half of our categories. We use soft label information from the source domain to provide 
information about the held-out categories which lack labeled target examples. 
Figure~\ref{fig:softlabel-res} examines one target example from the held-out category \emph{monitor}.
 No labeled target monitors were available during training; however, as shown in the upper right corner of Figure~\ref{fig:softlabel-res}, the soft labels for \emph{laptop computer} was present during training and  assigns a relatively high weight to the \emph{monitor} class. Soft label fine-tuning thus allows us to exploit the fact that these categories are similar. We see that the baseline model misclassifies this image as a \emph{ring binder}, while our soft label model correctly assigns the \emph{monitor} label.

\section{Conclusion}
We have presented a CNN architecture that effectively adapts to a new domain with limited or no labeled data per target category. We accomplish this through a novel CNN architecture which simultaneously optimizes for domain invariance, to facilitate domain transfer, while transferring task information between domains in the form of a cross entropy soft label loss. We demonstrate the ability of our architecture to improve adaptation performance in the \emph{supervised} and \emph{semi-supervised} settings by experimenting with two standard domain adaptation benchmark datasets.
In the semi-supervised adaptation setting, we see an average relative improvement of 13\% over the baselines on the four most challenging shifts in the Office dataset.
Overall, our method can be easily implemented as an alternative fine-tuning strategy when limited or no labeled data is available per category in the target domain.

\paragraph{Acknowledgements} This work was supported by DARPA; AFRL; DoD MURI award N000141110688; NSF awards 113629, IIS-1427425, and IIS-1212798; and the Berkeley Vision and Learning Center.

{\small
\bibliographystyle{ieee}
\bibliography{main}
}

\end{document}